%% file: sbi_main_paper.tex
\documentclass{article} 
\usepackage{iclr2023_conference_tinypaper,times}

\input{math_commands.tex}

\usepackage{graphicx}
\usepackage{hyperref}
\usepackage{url}
\usepackage{caption} 
\usepackage{adjustbox}

\title{CrossVoice: Crosslingual Prosody Preserving Cascade-S2ST  using Transfer Learning}

\author{Medha Hira\thanks{Equal Contribution},\hspace{2mm}Arnav Goel\footnotemark[1] \hspace{0.25mm} \& Anubha Gupta \\
Indraprastha Institute of Information Technology\\
New Delhi - 110020, India \\
\texttt{\{medha21265,arnav21519,anubha\}@iiitd.ac.in} \\
}

\iclrfinalcopy 
\begin{document}

\maketitle

\begin{abstract}


This paper presents CrossVoice, a novel cascade-based Speech-to-Speech Translation (S2ST) system employing advanced ASR, MT, and TTS technologies with cross-lingual prosody preservation through transfer learning. We conducted comprehensive experiments comparing CrossVoice with direct-S2ST systems, showing improved BLEU scores on tasks such as Fisher Es-En, VoxPopuli Fr-En and prosody preservation on benchmark datasets CVSS-T and IndicTTS. With an average  mean opinion score of 3.6 out of 4, speech synthesized by CrossVoice closely rivals human speech on the benchmark highlighting the efficacy of cascade-based systems and transfer learning in multilingual S2ST with prosody transfer.


\end{abstract}

\section{Introduction}
\vspace{-1em}
Transformer-based models \citep{vaswani2017attention} have revolutionized speech processing, leading to significant advancements in automatic speech recognition and text-to-speech technologies \citep{latif2023transformers, prabhavalkar2023endtoend}. This shift towards end-to-end systems has opened new avenues in Speech-to-Speech Translation (S2ST) for translating speech across languages. Our work introduces \emph{CrossVoice}, a cascade-based S2ST system utilizing the latest open-source automatic speech recognition (ASR), machine translation (MT), and text-to-speech (TTS) models unlike direct S2ST methods that bypass MT. It is evaluated against state-of-the-art (SOTA) direct S2ST systems for speech quality, cross-lingual prosody preservation, and translation accuracy using BLEU (BiLingual Evaluation Understudy) score \citep{papineni-etal-2002-bleu}. Further, we investigate the performance of cascade-based vis-\'a-vis direct approaches in S2ST and demonstrate how transfer learning can enhance prosody transfer in cross-lingual settings.
\vspace{-1em}
\section{Related Work} \label{related_work}
\vspace{-1em}
Current open-source systems for direct-S2ST involve various techniques such as self-supervised learning \citep{lee2021textless}, using speech discrete units \citep{lee2021direct}, text modalities \citep{zhang2023improving} and linguistic decoders \citep{jia2022translatotron}. However, these systems often face challenges including lower translation accuracy and inferior audio quality, particularly, in cross-lingual prosody transfer \citep{bentivogli2021cascade}. In contrast, cascade-based S2ST systems that integrate separate ASR, MT, and TTS models \citep{nakamura2006atr} are criticized for high latency and subpar prosody transfer \citep{latif-etal-2021-controlling}. 

Recent advancements in transfer-learning, such as voice cloning \citep{jia2019transfer} and transformer-based ASR and TTS, suggest the potential for more efficient and effective prosody transfer in cascade-based systems \citep{huang2023holistic}. Our study leverages these SOTA technologies in the proposed  cascade-based framework, \emph{CrossVoice}, and compares its performance with direct S2ST systems on prosody transfer and overall efficiency.

\begin{figure}
    \centering
\includegraphics[scale=0.25,trim=110 95 30 170]{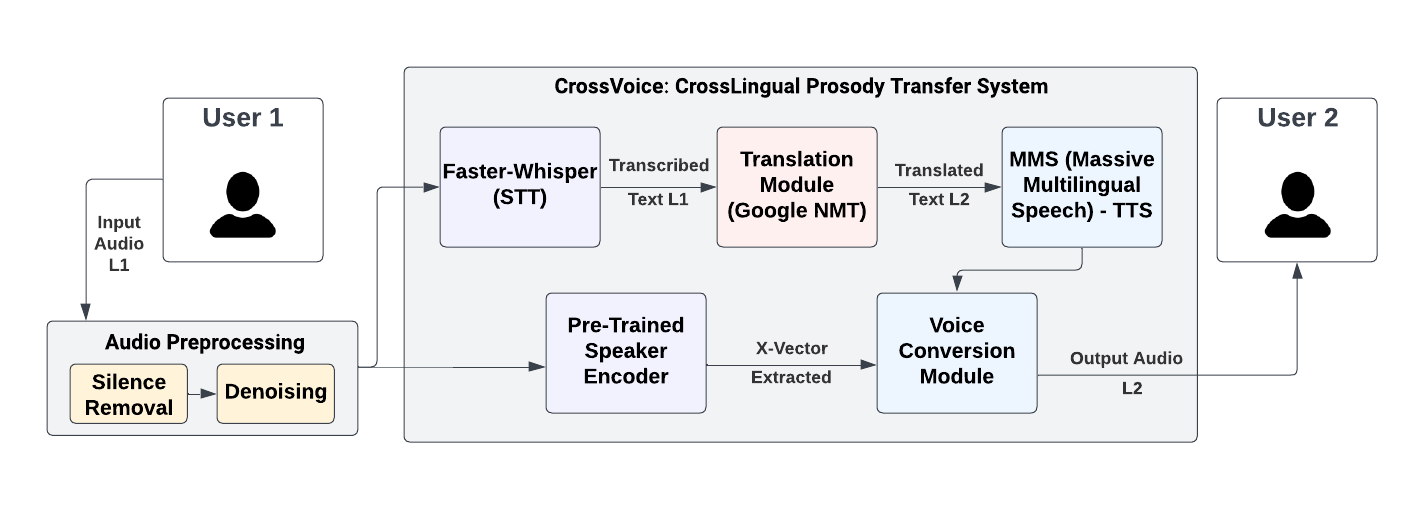}
    \caption{Proposed Architecture for CrossVoice}
    \label{fig:cross_voice}
\end{figure}
\vspace{-1em}
\section{Methodology}
\vspace{-1em}
CrossVoice integrates state-of-the-art ASR, MT, and TTS techniques to establish a baseline translation cascade: 1) Faster-Whisper\footnote{https://github.com/SYSTRAN/faster-whisper} for ASR {(comparision of other ASR models in \ref{a.2})}, which is a faster and batch-capable version of Whisper-Large  \citep{radford2022robust, moslem2022translation}; 2) Google's NMT model \citep{wu2016googles} for MT, which is known to reduce error rates significantly; and 3) the Massive Multilingual Speech (MMS) model \citep{pratap2023scaling} based on VITS-TTS \citep{kim2021conditional} for TTS, which is capable of handling over 1000 languages with superior performance in linguistic diversity and speech synthesis. CrossVoice uses transfer learning on a voice cloning module (trained on the speaker identification task) for prosody preservation. For this, a pre-trained speaker encoder generates X-vector embeddings \citep{speechbrain,snyder2018x} and is coupled with FreeVC's \citep{li2023freevc} voice conversion module to effectively transfer speaker prosody.

We conducted two sets of experiments to evaluate our system's performance in translation and speech synthesis. The first experiment evaluates synthesized speech quality on the CVSS-T \citep{jia2022cvss} and IndicTTS benchmark datasets \citep{kumar2023building}. We report mean opinion scores (MOS) from a survey of 40 respondents, rating on a four-point scale with a 95\% confidence interval as per the protocol of \cite{huang2023holistic}. MOS-h represents ratings for natural human speech (called as Ground Truth or GT here), MOS-v for baseline TTS audio without prosody transfer, and MOS-c for speech synthesized by CrossVoice\footnote{Details about MOS calculations and the protocol are given in the appendix \ref{a}}. The second experiment compares the BLEU performance of CrossVoice with recent direct-S2ST SOTA systems discussed in Section \ref{related_work} on the translation tasks for which their superiority has been claimed over cascade-based systems. 
\vspace{-1em}
\section{Results}
\label{methodology}
\vspace{-1em}

Table \ref{tab:translation_quality} tabulates results of the first experiment on five translation tasks. MOS-c score is almost the same as MOS-h (i.e., the GT) and also beats MOS-v scores of the vanilla TTS considerably, by almost 40\% on each task. Figures \ref{fig:X-en} and \ref{fig:en-X} (\textbf{see \ref{a.4}}) highlight high BLEU scores of CrossVoice that averaged to 33.4 over all the languages of the chosen benchmark datasets. 

Table-\ref{tab:s2s_metrics} lists the BLEU scores of CrossVoice (BLEU-c) and SOTA methods (BLEU-r). For calculating the BLEU scores, we employed Whisper (using the \emph{temperature setting of one} and \emph{greedy decoding}) for generating transcripts of the speech generated using CrossVoice and SOTA methods. We sourced BLEU scores from the original papers for the SOTA methods (reported as BLEU-r). CrossVoice surpasses the claimed superior performance of direct S2ST systems in their respective tasks, notably achieving almost a 19-point increase in BLEU score in the VoxPopuli S2ST Fr-En task. This significant performance boost is attributed to effective ASR and precise audio reconstruction through voice cloning.

\begin{minipage}{0.48\textwidth}
    \centering
 \captionof{table}{MOS comparison on S2ST quality} 
    \vspace{-1em}
\begin{adjustbox}{width=\textwidth,center}
    \begin{tabular}{|l|l|l|l|}
    \hline
    \small
    Translation Task & MOS-h ($\uparrow$)& MOS-v ($\uparrow$)$^\circ$& MOS-c ($\uparrow$)$^\circ$\\
     & (GT)& (Vanilla TTS) & (CrossVoice)\\
    \hline
    Spanish-English$^\dagger$   & 3.88 & 2.75 $\pm$ 0.12 & \textbf{3.76 $\pm$ 0.08}\\
    \hline
    German-English$^\dagger$   & 3.83 & 2.64 $\pm$ 0.05& 3.73 $\pm$ 0.11\\
    \hline
    Italian-English$^\dagger$ & 3.75 & 2.89 $\pm$ 0.01& 3.53 $\pm$ 0.10\\
    \hline
    Hindi-English$^\star$   & 3.79 & 2.54 $\pm$ 0.07& 3.63 $\pm$ 0.02\\
    \hline
    English-Hindi$^\star$    & 3.67 & 2.65 $\pm$ 0.03& 3.34 $\pm$ 0.04\\
    \hline
\end{tabular}
\end{adjustbox}
    \vspace{-1em}
\footnotetext{$^\dagger$CVSS-T, $^\star$Indic-TTS, $^\circ$mean$\pm$std}

\label{tab:translation_quality}
\end{minipage}%
\hfill 
\begin{minipage}{0.49\textwidth}
    \centering
    \captionof{table}{Comparison on S2ST-BLEU}
    \vspace{-1em}
    \begin{adjustbox}{width=\textwidth,center}
    \begin{tabular}{|l|l|l|}
        \hline
               \small
         Task (reported in & BLEU-r ($\uparrow$)& BLEU-c ($\uparrow$)\\ SOTA method)
          & (SOTA method) & (CrossVoice)\\
        \hline
        Fisher Es-En & 42.9 \citep{jia2022translatotron} & \textbf{45.6} \\ 
        \hline
        Fisher Es-En & 39.9 \citep{lee2021direct}& \textbf{45.6} \\
        \hline
        MuST-C En-De & 30.2 \citep{zhang2023improving}& \textbf{39.7}\\
        \hline
        MuST-C En-Fr & 40.8 \citep{zhang2023improving}& \textbf{46.5}\\
        \hline
        VoxPopuli Fr-En & 20.3 \citep{lee2021textless} & \textbf{39.6}\\
        \hline
    \end{tabular}
    \end{adjustbox}
    \vspace{3.5mm} 
    \label{tab:s2s_metrics}
\end{minipage}


%
\vspace{-1em}
\section{Conclusion and Future Work}
\vspace{-1em}
CrossVoice effectively combines advanced ASR, MT, and TTS technologies, establishing itself as a highly proficient cascade-based S2ST system with strengths in cross-lingual prosody preservation and translation accuracy. Our comprehensive experiments reveal that CrossVoice outperforms existing direct S2ST systems, underscoring the effectiveness and reliability of cascade-based systems with transfer learning for direct speech translation across languages. Future work includes improving transfer of emphasis and intonation across languages as reported in \ref{a.3}.

\subsubsection*{URM Statement}
The authors acknowledge that all the authors of this work meet the URM criteria of ICLR 2024 Tiny Papers Track.

\bibliography{sbi_main_paper.bib}
\bibliographystyle{iclr2023_conference_tinypaper}

\appendix
\section{Appendix} \label{a}

\subsection{Acronyms used}

\begin{itemize}
    \item MOS : Mean Opinion Score
    \item S2ST : Speech to Speech Translation
    \item ASR : Automated Speech Recognition
    \item MT : Machine Translation
    \item NMT : Neural Machine Translation
    \item TTS : Text to Speech
    \item GT : Ground Truth
    \item SOTA : State-Of-The-Art
    \item BLEU : Bilingual Evaluation Understudy
    
\end{itemize}

\subsection{ASR Results}\label{a.2}

We compared various ASR models such as variants of Whisper, Wav2Vec2.0 \citep{baevski2020wav2vec}, WavLM \citep{Chen_2022} and Faster-Whisper on multilingual datasets: Librispeech-test-clean (English), IndicTTS (Indian accented English speech) \citep{kumar2023building} and VoxPopuli - French, Spanish and German. Results are shown in Table \ref{table:model_comparison} and  \ref{table:model_comparison2}. Faster-Whisper clearly performs very well on both WER and average latency metrics. We measured average latency as the weighted average of the time taken to transcribe each sample of the entire dataset. 
\begin{table}[h]
\centering
\caption{Results of different ASR Models on Librspeech-test-clean subset \citep{panayotov2015librispeech}}
\vspace{-1em}
\begin{tabular}{|l|c|c|}
\hline
\textbf{Model} & \textbf{WER (\%)} & \textbf{Average Latency (s)} \\ \hline
Whisper - Tiny & 9.78 & 0.183 \\ \hline
Whisper - Base & 6.94 & 0.234 \\ \hline
Whisper - Small & 4.85 & 0.385 \\ \hline
Whisper - Large & 3.63 & 1.145 \\ \hline
Wav2Vec2.0 - Large & 3.20 & 0.415 \\ \hline
WavLM - Large & 2.80 & 0.525 \\ \hline
Faster-Whisper & \textbf{4.23} & \textbf{0.152} \\ \hline
\end{tabular}
\label{table:model_comparison}
\end{table}

\begin{table}[h]
\centering
\caption{WER benchmarking of models on various Datasets}
\vspace{-1em}
\begin{tabular}{|l|c|c|c|c|}
\hline
\textbf{Model} & \textbf{IndicTTS-en} & \textbf{VoxPopuli-French} & \textbf{VoxPopuli-Spanish} & \textbf{VoxPopuli-German} \\ \hline
Wav2Vec2.0 - XLSR & 15.65 & 25.34 & 21.34 & 24.73 \\ \hline
WavLM - Large & 14.25 & 23.21 & 18.65 & 20.56 \\ \hline
Whisper - Tiny & 10.74 & 31.53 & 19.63 & 25.24 \\ \hline
Whisper - Base & 8.63 & 21.34 & 15.32 & 19.75 \\ \hline
Whisper - Small & 5.28 & 13.24 & 12.18 & 13.32 \\ \hline
Whisper - Large & 3.85 & 10.56 & 7.82 & 9.75 \\ \hline
Faster-Whisper & 4.38 & 11.23 & 8.96 & 10.32 \\ \hline
\end{tabular}
\label{table:model_comparison2}
\end{table}

\subsection{Translation Tasks}
We benchmarked CrossVoice on 3 benchmark S2ST tasks and they are summarised as follows:
\begin{enumerate}
    \item Fisher (Spanish-English) \citep{post2014fisher}: The Fisher Spanish dataset is a collection of telephone speech conversations in Spanish, primarily involving topics of daily life. It contains over 160 hours of recorded conversations, involves more than 130,000 utterances and includes around 24,000 speakers.
    
    \item MuST-C (English to German \& English to French) \citep{di2019must}: It is a multilingual speech translation corpus with 273 hours of audio recorded for the English to German task and 236 hours of audio recorded for the English to French task. 

    \item VoxPopuli French-English \citep{wang2021voxpopuli}: French and English segments of the VoxPopuli dataset are taken for translation with 211 and 543 hours of transcribed audio. Same text segments from the dataset are taken for the S2ST task.
\end{enumerate}

For computing the BLEU-c scores, we randomly sampled 250 clips 10 times for each task and tested our system. The reported BLEU-c score is the average of these 10 iterations to ensure a fair and correct representation of our results.

\subsection{MOS Calculation Methodology and Protocol} \label{a.3}

Following the protocol laid out by \citep{huang2023holistic}, a survey was conducted of 40 respondents, where each respondent was shown the same set of translated clips along with their clips in the source language. This set consisted of 15 voice clips of duration varying from 2 secs to 10 secs. The following questions were asked from the respondents:
\begin{enumerate}
    \item Rate the similarity of the voice of the speaker to the original source clip : 1 - \emph{Completely Different}, 2 - \emph{Some similarities but more differences}, 3 - \emph{Some differences but more similarities}, 4 - \emph{Perfectly similar}.
    
    \item Rate the quality and naturalness of the generated audio clip: 1 - \emph{Extremely poor / robotic}, 2 - \emph{Somewhat natural but more robotic / poor}, 3 - \emph{Somewhat robotic/poor but more natural}, 4 - \emph{Perfectly natural}.

    \item Rate the similarity of the emphasis and intonation of the source clip and synthesised clip: 1 - \emph{Completely Different}, 2 - \emph{Some similarities but more differences}, 3 - \emph{Some differences but more similarities}, 4 - \emph{Perfectly similar}.
\end{enumerate}

Respondents were allowed to rate ``exactly in-between" for intermediary cases. It was noted starkly that on the first two questions, a huge proportion of respondents rated the synthesised speech for the five languages as close to 4. However, on the last question, a lot of respondents rated the system between 2 and 3 indicating that while the speaker's voice characteristics and prosody are being transferred with quality, intonation and emphasis will need improvement.


For calculating MOS-v, we employed our MMS TTS without using any voice cloning. Similar surveys on a lesser number of clips were able to see the Vanilla TTS system getting lower ratings compared to CrossVoice on all the three questions. We referenced MOS-h scores from the official paper of \citep{jia2022cvss}. 

\subsection{Results on CVSS-T and IndicTTS}\label{a.4}
\begin{figure} []
    \centering
    \includegraphics[width=1\linewidth]{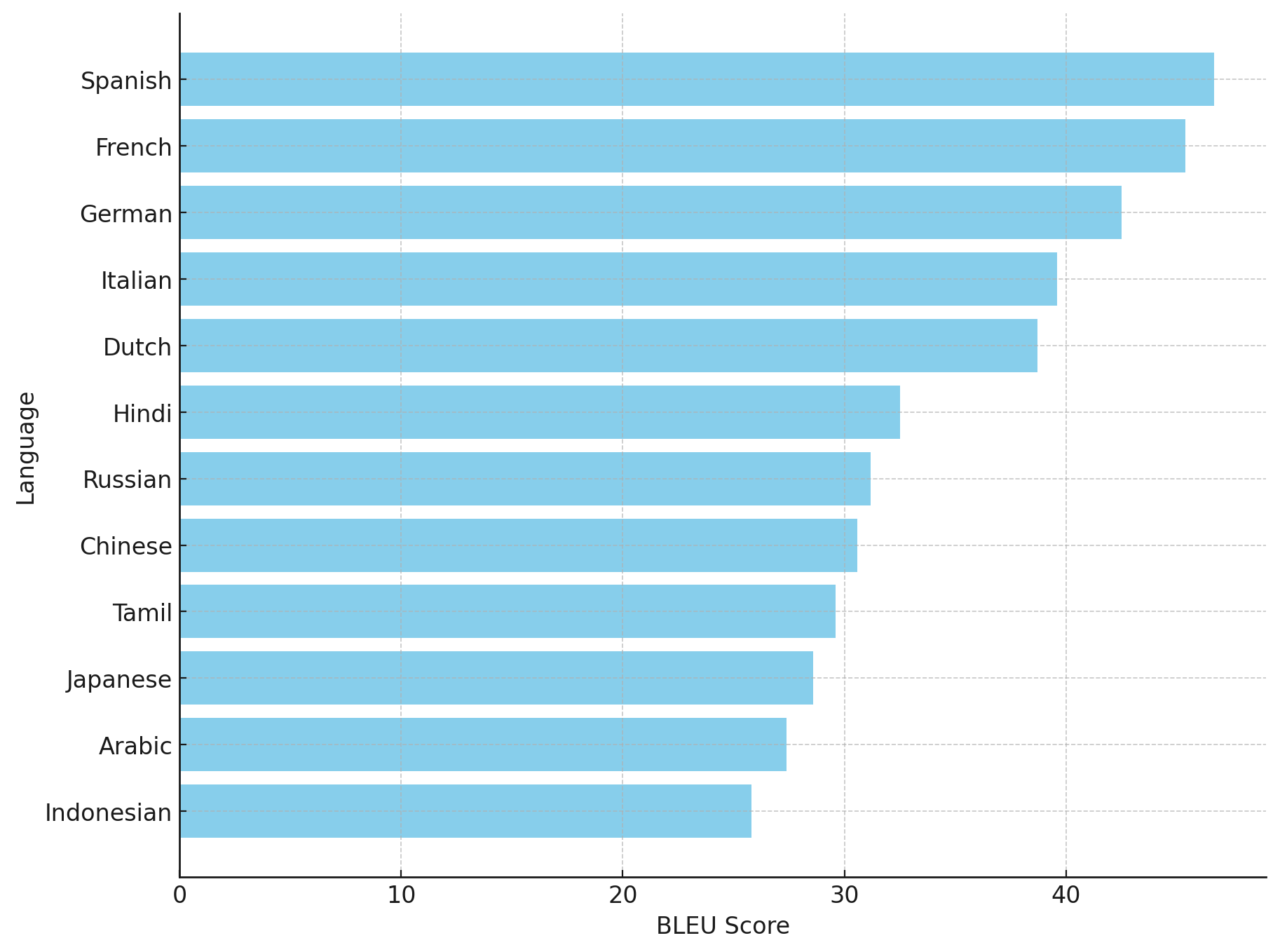}
    \caption{BLEU scores on CVSS-T for X-en Translation}
    \label{fig:X-en}
\end{figure}

\begin{figure}
    \centering
    \includegraphics[width=1\linewidth]{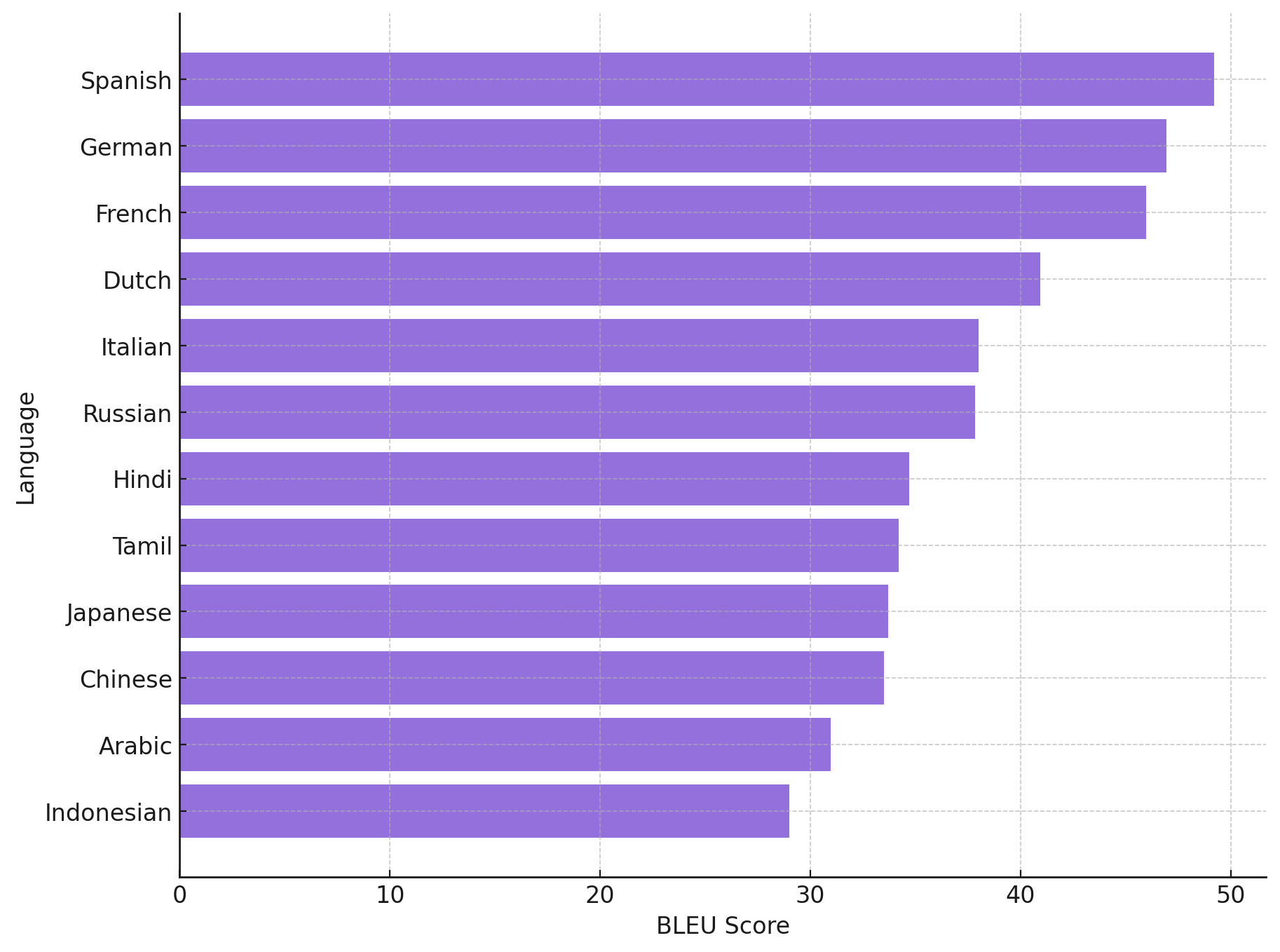}
    \caption{BLEU scores on CVSS-T for en-X Translation}
    \label{fig:en-X}
\end{figure}

We conducted experiments on 11 languages from the CVSS-T dataset and Hindi from IndicTTS dataset using CrossVoice. Figure \ref{fig:X-en} shows the results for these 12 languages when translated from any language $X \longrightarrow en$ (English), whereas Figure \ref{fig:en-X} shows the results for the 12 languages when translated from (English) $en \longrightarrow X$. Notably, our system shows higher BLEU scores on translating from English to any language because of low WER of Whisper on English and NMT being extensively pre-trained on $en \longrightarrow X$ tasks.

For calculating the results, we randomly took samples of 100 clips for each language and calculated results for one sample. We repeated this process for 10 iterations to check for biases. We report the average BLEU score for each language from the experiments. The standard deviation shown on all the tasks ranged between $\in (0.5, 1.5)$, thus, indicating lesser deviation.

\subsection{Ethical Considerations}
This study has been conducted and tested on standard open source datasets (that are appropriately cited in the paper), widely used in the literature. 

We recognize that voice cloning has the potential to be used for malicious activities; however, the benefits of this technology may outweigh the negatives. Our system is designed to encourage inclusivity and transcend the language barrier in communication between individuals. 

Further, we advocate for transparency in the use of voice cloning technology and users should always be informed when they are interacting with a cloned voice.

\subsection{Limitations and Challenges}
{CrossVoice relies heavily on extensive datasets for training. Obtaining and processing large, high-quality, and diverse datasets that cover a wide range of languages and accents is a significant challenge and can limit the system's effectiveness and scalability. CrossVoice encounters challenges in accurately transferring prosody, like intonation and stress patterns, across different languages. This is a complex task due to the inherent differences in linguistic structures and prosodic features among languages. This lack of appropriate transfer of intonation and emphasis is also depicted by the MOS score protocol \ref{a.3}}
\end{document}

%% file: math_commands.tex
\usepackage{amsmath,amsfonts,bm}

\def\eqref#1{equation~\ref{#1}}

\def\1{\bm{1}}

\DeclareMathAlphabet{\mathsfit}{\encodingdefault}{\sfdefault}{m}{sl}
\SetMathAlphabet{\mathsfit}{bold}{\encodingdefault}{\sfdefault}{bx}{n}

